\documentclass[journal]{IEEEtran}

\usepackage{hyperref}
\usepackage{amsmath,amssymb,amsfonts}
\usepackage{bm}
\usepackage{graphicx}
\usepackage{textcomp}
\usepackage{subfigure}
\usepackage[misc]{ifsym}
\usepackage{color}
\usepackage{booktabs}
\usepackage{multirow}
\usepackage{times}
\usepackage{float}
\usepackage{epsfig}
\usepackage{epstopdf}
\usepackage{hyperref}
\usepackage{mathrsfs,amsmath}
\usepackage{subfigure}
\usepackage[switch]{lineno}
\usepackage[ruled,vlined]{algorithm2e}

\usepackage[normalem]{ulem}

\begin{document}
%
\title{Dynamic Slimmable Denoising Network}

\author{Zutao Jiang,
       Changlin Li,
       Xiaojun Chang,~\IEEEmembership{Senior Member,~IEEE }
       Jihua Zhu,
       and Yi Yang,~\IEEEmembership{Senior Member,~IEEE }
\thanks{Z. Jiang and J. Zhu are with the School of Software Engineering, Xi’an Jiaotong Universiy, P. R. China. (E-mail: taozujiang@gmail.com; zhujh@xjtu.edu.cn).}
\thanks{C. Li is with Department of Data Science and AI, Monash Univeristy.}
\thanks{X. Chang is with the School of Computing Technologies, RMIT University, Australia. (E-mail: cxj273@gmail.com).}
\thanks{Yi Yang is with the Australian Artificial Intelligence
Institute, University of Technology Sydney, Ultimo, NSW 2007, Australia. (E-mail: yi.yang@uts.edu.au)}
\thanks{Corresponding author: Jihua Zhu.}
}

\markboth{IEEE Transactions on Image Processing}%
{Jiang \MakeLowercase{\textit{et al.}}: DDS-Net: Denoising via Dynamic Slimmable  Network}

\maketitle

\begin{abstract}
Recently, tremendous human-designed and automatically searched neural networks have been applied to image denoising. However, previous works intend to handle all noisy images in a pre-defined static network architecture, which inevitably leads to high computational complexity for good denoising quality. Here, we present dynamic slimmable denoising network (DDS-Net), a general method to achieve good denoising quality with less computational complexity, via dynamically adjusting the channel configurations of networks at test time with respect to different noisy images. Our DDS-Net is empowered with the ability of dynamic inference by a dynamic gate, which can predictively adjust the channel configuration of networks with negligible extra computation cost. To ensure the performance of each candidate sub-network and the fairness of the dynamic gate, we propose a three-stage optimization scheme. In the first stage, we train a weight-shared slimmable super network. In the second stage, we evaluate the trained slimmable super network in an iterative way and progressively tailor the channel numbers of each layer with minimal denoising quality drop. By a single pass, we can obtain several sub-networks with good performance under different channel configurations. In the last stage, we identify easy and hard samples in an online way and train a dynamic gate to predictively select the corresponding sub-network with respect to different noisy images. Extensive experiments demonstrate our DDS-Net consistently outperforms the state-of-the-art individually trained static denoising networks. 
\end{abstract}
\IEEEpeerreviewmaketitle

\section{Introduction}
\IEEEPARstart{I}{mage} denoising, which aims at recovering clean images from corrupted noisy images, is one of the most fundamental and challenging tasks in the low-level computer vision field. Recently, deep neural networks have been widely used in the field of image denoising. To achieve higher denoising results, these neural networks are growing deeper and wider, leading to higher computational complexities. Due to the high computational complexities of these heavy deep denoisers, it is impractical to deploy them in mobile platforms, such as mobile phones, mobile robots, autonomous cars and augmented reality devices. Therefore, there is an urgent need to explore efficient denoising models. 

In recent years, efficient network design \cite{howard2017mobilenets, sandler2018mobilenetv2, tan2019efficientnet}, network pruning \cite{he2018soft, he2017channel, liu2019metapruning}, knowledge distillation \cite{hinton2015distilling}, weight quantization \cite{jacob2018quantization} and dynamic neural network \cite{hua2018channel, veit2018convolutional, wang2018skipnet, li2021dynamic} have played an important role in improving computational efficiency of neural network. However, among these techniques, only static efficient network design has been used for image denoising \cite{gu2019self, xu2021efficient}, while most of other techniques are commonly used in the filed of high-level computer vision. Besides, these two methods on efficient denoising \cite{gu2019self, xu2021efficient} only evaluated on additive Gaussian denoising benchmarks and fail to handle real-world noisy images.

In this paper, we attempt to develop an efficient and accurate method for image denoising via instance-wise dynamic neural network, which aims to process different inputs with data-dependent parameters or architectures \cite{han2021dynamic}. Given different inputs, networks with dynamic architectures can adaptively perform inference with dynamic depth \cite{huang2017multi}, width \cite{hua2018channel, yuan2020s2dnas} or routing \cite{tanno2019adaptive} within a super network, resulting in remarkable advantages with regard to accuracy and computational efficiency. However, it is observed that there is an enormous gap between the practical acceleration of dynamic networks and the theoretical efficiency \cite{han2021dynamic}. For instance, some dynamic networks involving sparse computation \cite{he2017channel, liu2019metapruning} are inefficient on contemporary GPUs. This is because that the sparse computation pattern is not compatible with the current hardware and libraries, which are originally developed for static deep learning models. Therefore, it is important and worthwhile to design dynamic networks that are compatible with existing hardware, so as to achieve actual acceleration instead of only theoretical efficiency. 

Recently, Yu et al. \cite{yu2018slimmable} proposed slimmable neural networks (S-Net), which allows adaptive trade off between accuracy and efficiency via adjusting the width of neural networks on the fly. The width of neural networks is chosen from a predefined widths set and the parameters of slimmable networks with different width are shared. To make slimmable networks capable of executing at arbitrary width, Yu et al. \cite{Yu2019UniversallySN} then proposed a universally slimmable networks (US-Net). In US-Net, the value of a single output neuron is an partial or full aggregation of weighted input neurons $y = \sum\nolimits_{i = 1}^{\left\lceil {\rho n} \right\rceil } {{w_i}{x_i}}$, where $x$ is input neuron, $y$ is output neuron, $w$ is the convolution filter. $\rho$ is network width ratio, (ranges from 0 to 1), $n$ is the number of input channels and $\left\lceil  \cdot  \right\rceil $ denotes the ceiling operation. As US-Net can keep filters static and contiguous when switching the width of networks, US-Net can achieve actual acceleration.

\begin{figure}[!t]
\centerline{\includegraphics[width=0.5\textwidth]{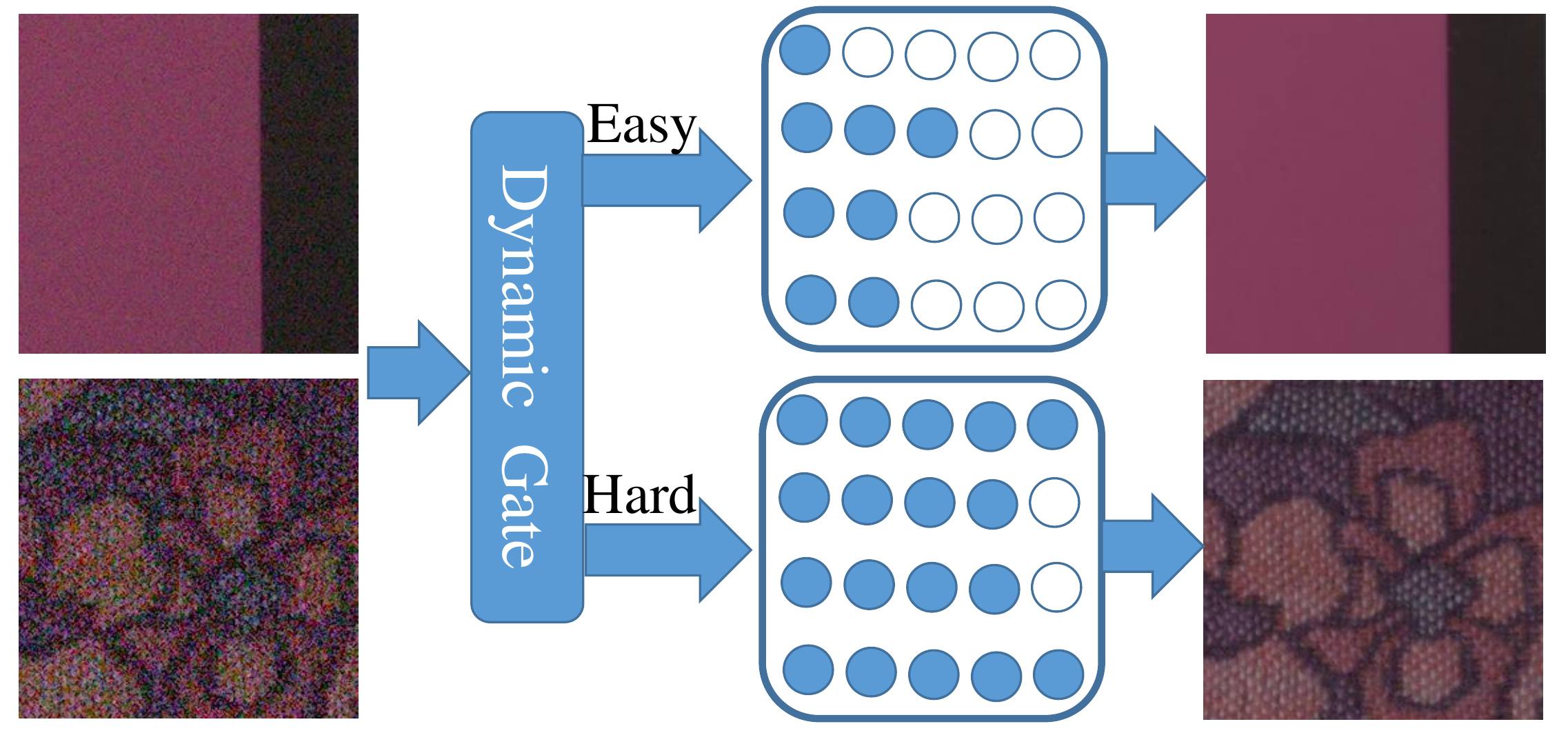}}
\caption{Conditioned on the input noisy images, DDS-Net activates different numbers of neurons of the slimmable super network. Here, each row of circles is a layer of the neural network. The solid circles represents the activated neurons and the hollow circles denotes the inactivate neurons.}
\label{fig:hard_easy}
\end{figure}

Inspired by \cite{li2021dynamic, Yu2019UniversallySN}, we propose an efficient and accurate deep image denoising method via dynamic slimmable network, named DDS-Net. DDS-Net can achieve an adaptive trade off between denoising performance and efficiency by activating different numbers of channels of slimmable network conditioned on different noisy images. Similar to US-Net, modern GPUs and deep learning libraries are also friendly to DDS-Net. As illustrated in Fig. \ref{fig:hard_easy}, we use a dynamic gate to make data-dependent decisions for the channel configuration of slimmable networks during inference. As the optimization of slimmable super network and the dynamic gate is strongly coupled, we propose a three-stage optimization scheme for DDS-Net. 

In the first stage, we use the Sandwich Rule \cite {Yu2019UniversallySN} to train a weight-shared slimmable super network. In each training iteration, we train the denoising models with different channel configurations all together, including the smallest model, the largest model and other models with randomly sampled channel configurations. To further improve the performance of the super network, we propose In-place Synergy which transfers the knowledge inside a single super network from the larger sub-networks to the smaller sub-networks. In the second stage, inspired by [21], we present a progressive slimming technique to optimize our dynamic routing space in static way before dynamic adaptation. The trained slimmable model are evaluated in an iterative way and the channels of each layer are greedily 
trimmed with minimal performance drop. By a single pass, we can acquire several sub-networks with the optimized channel configuration. In the last stage, we first identify the difficulty level of the input noisy images according to the performance gap between the smallest sub-network and the largest sub-network, and then train the dynamic gate. With the trained dynamic gate, the smaller sub-networks are used for easy inputs while larger sub-networks tend to handle hard inputs. 

Overall, our contributions can be summarized as follows:
\begin{itemize}
\item We propose an efficient and accurate deep image denoising method via dynamic slimmable network, which can achieve good trade off between denoising quality and efficiency. The proposed method can achieve actual acceleration because the weight parameters of each layer are static and contiguous when the sub-networks switch.
\item We propose a three-stage optimization scheme which can decouple the training of slimmable super network, optimization of the architecture and dynamic gate. With the proposed three-stage optimization scheme, we can obtain a super network with optimized channel configurations and a fair gate network.
\item We use In-place Synergy technique to assist the training of slimmable super network, which can transfer the knowledge inside a single super network form the larger sub-network to the smaller sub-network.
\item Extensive experiments demonstrate that our DDS-Net consistently outperforms the state-of-the-art individually trained static denoising network.
\end{itemize}

\section{Related Works} \label{Rw}
In this section, we mainly discuss previous methods on CNN-based image denoising, dynamic neural networks, slimmable networks and knowledge distillation which are close to this work.

\subsection{CNN-based image denoising}
With the rapid development of deep neural networks, CNN-based image denoising algorithms are becoming prevalent and have achieved a significant performance boost. As the early attempt, Jain el al. \cite{jain2008natural} applied a simple CNN for image denoising and obtained comparable performance with the wavelet and Markov random field (MRF) methods. To boost the denoising performance, Zhang et al. \cite{zhang2017beyond} incorporated residual learning and batch normalization into deep CNN. To make the training of deep CNN easier and more effective, Mao et al. \cite{mao2016image} introduced symmetric skip connections into 30-layer convolutional encoder-decoder networks for image denoising. To handle the long-term dependency problem in deep CNN architectures, \cite{tai2017memnet} proposed a end-to-end persistent memory network (MemNet), which use a memory block to adaptively bridge the long-term dependencies. To make use of the non-local self-similarity in natural images, Liu et al. \cite{liu2018non} combined recurrent neural network with non-local operation, resulting in a non-local recurrent network (NLRN). To obtain a better trade off between receptive field size and computational efficiency, Liu et al. \cite{liu2018multi} presented a multi-level wavelet CNN model (MWCNN), which is comprised of a contracting sub-network and a expanding sub-network. To generate high quality denoising results with less runtime and GPU memory, Gu et al. \cite{gu2019self} proposed a self-guided neural network (SGN), which can utilize multi-scale contextual information via a top-down self-guided architecture. Instead of down-samping or pooling operation, SGN adopts shuffling operation to generate multi-resolution inputs, which can avoid information loss and achieve a better denoising performance. Similarly, Xu et al. \cite{xu2021efficient} recently proposes an efficient deep neural network for image denoising based on pixel-wise classification. They replace part of the convolution layers in existing denoising networks by Class Specific Convolution layers (CSConv) which use different weights for different classes of pixels. Most of the denoising methods mentioned above are aimed at dealing with Gaussian denoising task.

In recent years, there have been some works focusing on real-word noisy image denoising. In \cite{guo2019toward}, CBDNet was proposed for real-world noisy image denoising, by considering in-camera signal processing pipeline. In \cite{anwar2019real}, RIDNet employed feature attention in real-world image denoising, so as to exploit the channel dependencies. In \cite{yue2019variational}, VDNet integrated image denoising and noise estimation into a unique Bayesian framework. In \cite{yue2020dual}, DANet combined the noise generation and the noise removal tasks into a unified framework. DANet is composed of a denoiser and a generator. The learned generator can be used to augment the original training data, leading to the high denoising performance of the denoiser. In \cite{zamir2020learning}, MIRNet developed a multi-scale residual block that can extract enriched multi-scale contextual features and preserve the precise spatial details. In \cite{kim2020transfer}, Kim et al. incorporated transfer learning in the real-world image denoising, which is capable of transferring knowledge learned from synthetic-noise training data to the real-noise denoiser. In \cite{Zamir2021MPRNet}, Zamir et al. proposed a multi-stage architecture for image denoising, which can progressively recovery clean images form degraded inputs. However, one of the limitation of the above works for real-word image denoising is the heavy computational consumption. Compared to the works mentioned above, our DDS-Net can achieve good performance with less computational complexity for the real-world image denoising.

\subsection{Dynamic neural networks}
Dynamic neural networks adjust their architectures or parameters conditioned on the inputs. In contrast to static models, dynamic neural networks have notable advantages with regard to performance and computational efficiency. Generally, dynamic networks fall into three categories: instance-wise dynamic networks, spatial-wise dynamic networks and temporal-wise dynamic networks. In this paper, we mainly emphasis on instance-wise dynamic networks.

Instance-wise dynamic networks process disparate inputs with different model architectures, therefore improving the computation efficiency. To reduce the average inference time of deep neural networks, Bolukbasi et al. \cite{bolukbasi2017adaptive} presented an adaptive early-exit strategy which enables easy inputs to bypass some layers of networks. To solve the dynamic skipping problem, Wang et al. \cite{wang2018skipnet} proposed a modified residual network, named SkipNet. In SkipNet, convolutional blocks are selectively skipped via a gating network. To design dynamic networks along the width (channel) dimension, Hua et al.\cite{hua2018channel} introduced a channel gating neural network (CGNet), which utilizes an activation-wise gate function to identify effective input channels and performs the convolution operation on them. Similarly, Lin et a. \cite{lin2017runtime} developed a Runtime Neural Pruning (RNP) network, which conducts channel-wise dynamic pruning with a Markov decision process. Instead of using a gate function or a policy network, Liu et al. \cite{liu2019learning} introduced structural feature sparse regularization on deep CNN models and developed an instance-wise feature pruning. To achieve dynamic routing, Tanno et al. \cite{tanno2019adaptive} incorporated decision tree into deep neural networks and developed the adaptive neural trees (ANTs). More recently, Li et al. \cite{li2021dynamic} proposed a new dynamic network routing regime with a two-stage training framework, which can achieve good hardware-efficiency. There are two main differences between our DDS-Net and \cite{li2021dynamic}: 1) our DDS-Net focuses on image denoising while \cite{li2021dynamic} is applied on image classification; 2) we propose a three-stage optimization scheme when \cite{li2021dynamic} use a two-stage training framework.

\subsection{Slimmable networks}
Slimmable networks are a group of parameter-shared neural networks executable at different widths. In \cite{yu2018slimmable}, Yu et al. first developed switchable slimmalbe networks (S-Net) which allows instant and adaptive trade off between accuracy and efficiency at inference time. The widths of S-Net are chosen form a predefined widths list. Based on S-Net, Yu et al. \cite{Yu2019UniversallySN} then proposed universally slimmalbe networks (US-Net), which can run at arbitrary width. To boost testing performance, US-Net are trained with two techniques: the sandwich rule and in-place distillation. Instead of setting channel numbers in a neural network based on heuristics, Yu et al. \cite{yu2019autoslim} presented a one-shot method, namded AutoSlim, which can automatically search optimal channel configurations under constrained resource.

\subsection{Knowledge distillation}
Knowledge distillation has been applied to a wide range of fields including computer vision, speech recognition, network compression and so on. The basic idea of knowledge distillation is to transfer knowledge acquired from a cumbersome network to a compact network, leading to the performance improvement for the compact network. The concept of knowledge distillation in deep neural networks was first introduced by Hition et al \cite{hinton2015distilling}. Generally, soft targets \cite{hinton2015distilling} or intermediate feature \cite{komodakis2017paying} from the cumbersome model can be used to train the compact model. Besides, the compact network can learn better knowledge from the ensemble of multiple cumbersome networks, which is more informative \cite{tarvainen2017mean, you2017learning}. 

\section{Method}
\label{Sec:Method}
The goal of our method is to reduce the computational complexity of existing state-of-the-art deep learning-based denoisers, while maintaining the denoising performance. Towards this end, we propose a deep image denoising method via dynamic slimmable network, named as DDS-Net. Given the noisy images, DDS-Net allocates appropriate computation by adjusting the channel configuration of neural networks, so as to reduce the redundant computation on those easy noisy images. 

Formally, given a noisy image $\bm y$ and the corresponding ground-truth image $\bm x$. The objective of dynamic slimmable denoising network is to choose an optimal architecture $\bm\alpha^\star(\bm y)$ with respect to the noisy image $\bm y$:
\begin{equation}
\begin{aligned}
    &\bm\alpha^\star(\bm y) =\mathop{\arg\min}\limits_{\bm\alpha\in\mathcal{A}}  \mathcal{L}\Big(\mathcal{F}\big(\mathbf{\Theta}^\star, \bm\alpha(\bm y); \bm{y}\big), \bm x\Big),\\
    s.t. ~~&\mathbf{\Theta}^\star = \mathop{\arg\min}\limits_\mathbf{\Theta}  \mathcal{L}\Big(\mathcal{F}\big(\mathbf{\Theta}, \bm\alpha^\star(\bm y); \bm{y}\big), \bm x\Big),\\
\end{aligned}
\label{eq:dyn_obj}
\end{equation}
where $\mathcal{F}\big(\mathbf{\Theta}, \bm\alpha(\bm y); \bm{y}\big)$ is a weight shared super network with parameter $\mathbf{\Theta}$ and routing space $\mathcal{A}$. $\bm\alpha(\bm y)$ is a single route chosen from the routing space $\mathcal{A}$ with respect to the noisy image $\bm y$. 
As the optimization of slimmable super network and the dynamic gate is strongly coupled, we propose a three-stage optimization scheme for DDS-Net to decouple the optimization of the super network parameters $\mathbf{\Theta}$, routing space $\mathcal{A}$ and the dynamic architecture $\bm\alpha$:
\begin{equation}
\left\{
\begin{aligned}
    &\mathbf{\Theta}^\star = \mathop{\arg\min}\limits_\mathbf{\Theta}  \sum\limits_{\bm{\alpha}\in\mathcal{A}}\mathcal{L}\big(\mathcal{F}(\mathbf{\Theta}, \bm \alpha; \bm{y}), \bm x\big),\\
    &\mathcal{A}^\star =\Big\{\mathop{\arg\max}\limits_{\bm \alpha \in \widetilde{\mathcal{A}}_i}  Acc\big(\mathcal{F}(\mathbf{\Theta}^\star, \bm \alpha)\big)\Big\}_{i=0}^N, \\
    &\bm{\varphi}^\star = \mathop {\arg \min }\limits_{\bm\varphi} \mathcal{L}\big(\mathcal{G}(\bm{\varphi}; \bm y), \mathbf{G}(\bm y)\big) +{\left( {\frac{{\mathbf{\Psi}(\bm{\varphi}; \bm y ) - C}}{T}} \right)^2},\\
\end{aligned}
\right.
\label{eq:three_stage_opt}
\end{equation}
where $\widetilde{\mathcal{A}}_i$ is the set of sub-networks for the $i$-th iteration of the progressive slimming, $N$ is the number of iterations during the progressive slimming. $\mathcal{G}(\bm{\varphi}; \bm y)$ is a dynamic gate network with parameter $\bm{\varphi}$. $\mathbf{G}(\bm y)$ represents the gate label for the noisy image $\bm y$. $\mathbf{\Psi}(\bm{\varphi}; \bm y )$ is the computation cost for recovering $\bm y$, $C$ is the computation constraint of hardware, $T$ denotes the normalization factor. After the three-stage optimization, the optimal route for each input can be derived from dynamic inference in the optimal routing space $\mathcal{A}^\star$ with the gate network $\mathcal{G}(\bm{\varphi}^\star; \bm y)$:
\begin{equation}
\bm\alpha^\star(\bm y) = \mathcal{G}(\bm{\varphi}^\star; \bm y) \cdot \mathcal{A}^\star.
\end{equation}

\begin{figure}[!t]
\centerline{\includegraphics[width=0.5\textwidth]{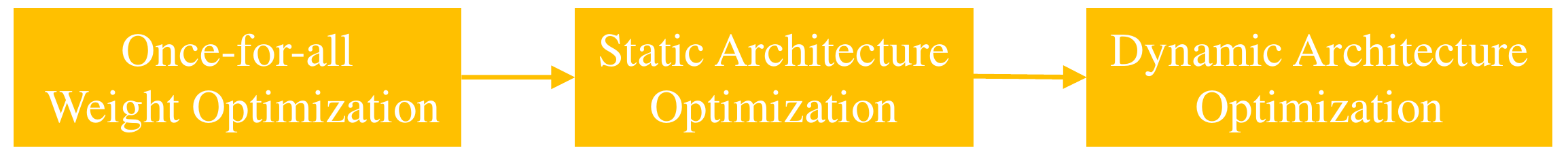}}
\caption{The three-stage optimization pipeline of DDS-Net.}
\label{fig:three_stage}
\end{figure}

The optimization pipeline of DDS-Net is displayed in Fig. \ref{fig:three_stage}. As shown in Fig. \ref{fig:three_stage}, the optimization of DDS-Net is composed of three stages: once-for-all weight optimization via In-place Synergy, static architecture optimization via progressive slimming, dynamic architecture optimization via dynamic gate. \textbf{Algorithm \ref{tab:algorithm}} summarizes the three-stage optimization of DDS-Net. In the following part of this section, we will present them in details.

\begin{algorithm}[ht]
\footnotesize

\SetAlgoLined
\textbf{Input:}\\  
$\mathcal{A}$: routing space;\\
$\bm\Theta$: super network parameters;\\
$\mathcal{D}$: training set;\\
$T_s$: number of epochs for the super network training;\\
$T_g$: number of epochs for the gate function training;\\
$L$: the number of layers of the super network;\\
\textbf{Stage 1: Once-for-all Weight Optimization} \\
\For{$i = 1,2, ..., T_s$}{
    \For{$\bm x, \bm y\in\mathcal{D}$}{
    Sample sub-networks $\{\alpha\} \subset \mathcal{A}$ following \textit{sandwich rule};\\
    Compute \textit{In-place Synergy} loss with $\{\alpha\},\bm\Theta, \bm x, \bm y$; \\ 
    Compute gradients and update $\bm\Theta$;\\
    }
}
Save the final $\bm\Theta$ as $\bm\Theta^\star$;\\

\textbf{Stage 2: Static Architecture Optimization} \\

Initialize $\mathcal{A}^\star$ as $\emptyset$;\\
$\alpha \leftarrow \alpha_\textit{largest}$;\\
\While{$\alpha$ is not $\alpha_\textit{smallest}$}{
    \For {$\ell = 1, 2, ..., L$}{
        Get $\alpha_\ell$ by slimming a group of channels in layer $\ell$ of $\alpha$;\\
        Evaluate $\alpha_\ell$ with $\bm\Theta^\star$;\\ 
        If performance is optimal, save $\ell$ as $\ell^\star$;\\
    }
    
    Slim $\alpha$ in the layer $\ell^\star$ by $\alpha\leftarrow\alpha_{\ell^\star}$;\\ 
    $\mathcal{A}^\star\leftarrow\mathcal{A}^\star \cup \{\alpha_{\ell^\star}\}$;\\
    
}

\textbf{Stage 3: Dynamic Architecture Optimization} \\
\For{$i = 1, 2, ..., T_g$}{
    \For{$\bm x, \bm y\in\mathcal{D}$}{
        Generate online gate label $\mathbf{G}(\bm y)$ by Eq. \ref{Eq:generate_online_label}; \\
        Calculate gate loss by Eq. \ref{Eq:gate_loss};\\ 
        Compute gradients and update $\bm\varphi$;\\
    }
}
Save the final $\bm\varphi$ as $\bm\varphi^\star$;\\
 \KwResult{
 $\bm\Theta^\star$, $\mathcal{A}^\star$, $\bm\varphi^\star$: optimal super network parameters, optimal routing space and optimal gate function parameters.}
\caption{Three-stage Optimization of DDS-Net}\label{tab:algorithm}
\end{algorithm}

\subsection{Once-for-all Weight Optimization via In-place Synergy}
Current deep neural networks are made up of several layers and each layer is comprised of a number of neurons. Generally speaking, the value of output neuron is the weighted sum of all input neurons. Mathematically, The calculation of output neuron value is commonly formulated as:
\begin{equation}
\bm y = \sum\nolimits_{i = 1}^n {{w_i}{x_i}}
\end{equation}
where $\bm x = \left\{ {{x_1},{x_2}, \cdots ,{x_n}} \right\}$ is the input neurons, $\bm w = \left\{ {{w_1},{w_2}, \cdots ,{w_n}} \right\}$ is the weights of input neurons, $\bm y$ is the output neuron, $n$ is the number of input neurons. Specifically, $n$ is the number of channels for the convolutional neural networks (CNNs). It is widely accepted that different neurons are in charge of detecting different features and the number of neurons has a significant impact on the performance and efficiency of denoising model. However, in static CNNs, the number of channels is manually set based on heuristic and models with different channel configurations need to be trained individually. To adjust the channel configuration of deep model conditioned on the input noisy images, we first need to train a weight-shared flexible super network which can executable with arbitrary number of channels. 

Let $\rho$ be the width multiplier of a neural network layer, the calculation of output neuron value can be rewritten as follows:
\begin{equation}
\bm y = \sum\nolimits_{i = 1}^{\left\lceil {\rho n} \right\rceil } {{w_i}{x_i}},    
\end{equation}
where $\bm x = \left\{ {{x_1},{x_2}, \cdots ,{x_{\left\lceil {\rho n} \right\rceil }}} \right\}$ is the input neurons, $\bm w = \left\{ {{w_1},{w_2}, \cdots ,{w_{_{\left\lceil {\rho n} \right\rceil }}}} \right\}$ is the corresponding weights of input neurons, $\bm y$ is the output neuron, $\left\lceil  \cdot  \right\rceil $ represents the ceiling operation, $\left\lceil {\rho n} \right\rceil $ denotes the number of input neurons. One native solution is to accumulate back-propagated gradients of sub-networks with all channel configurations. However, as there are tons of sub-networks with different channel configurations in the super network, it is not feasible to accumulate gradients of all sub-networks in each training iteration. 

\begin{figure}[!t]
\centerline{\includegraphics[width=0.4\textwidth]{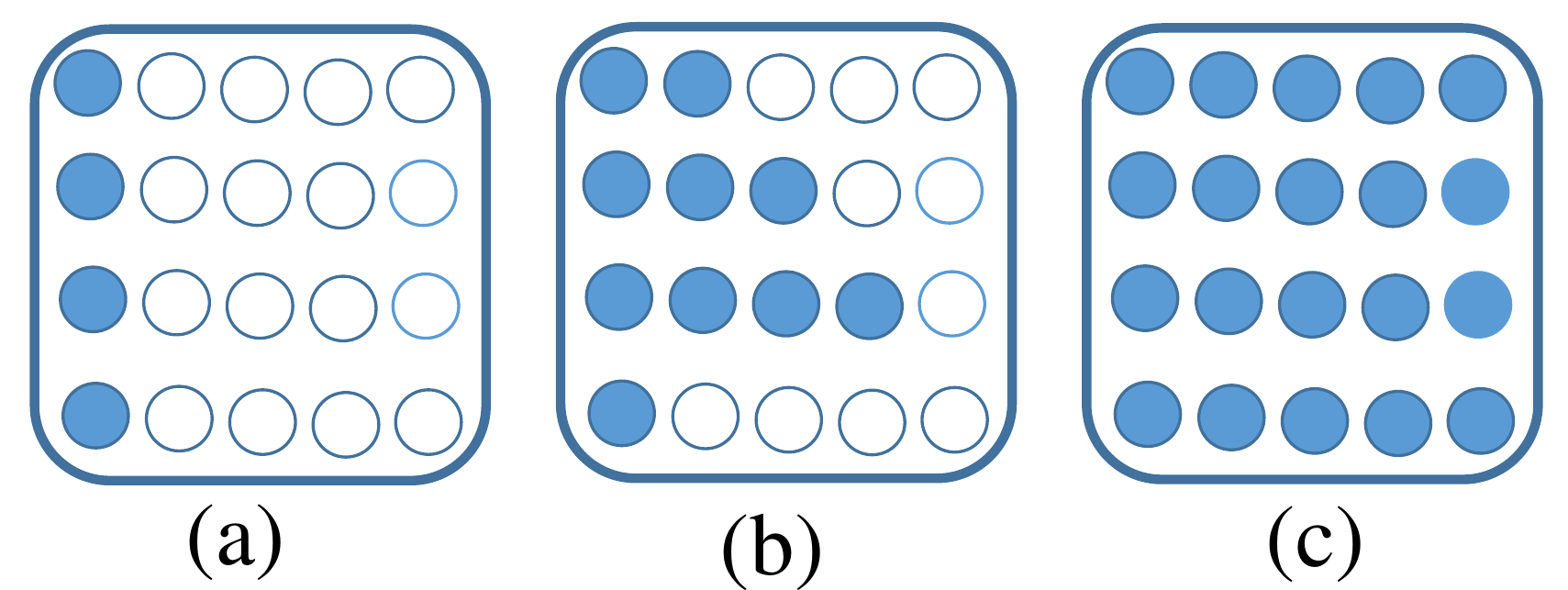}}
\caption{Three types of sub-networks used for super network training. (a) The smallest sub-network, (b) The randomly sampled sub-network, (c) the largest sub-network. All these sub-networks are sampled from a weight-shared super-network. Each row of circles is a layer of the neural network. The solid circles represents the activated channels and the hollow circles denotes the inactivate channels.}
\label{fig:sub-networks}
\end{figure}

In order to train the weight-shared super network efficiently, the sandwich rule \cite{Yu2019UniversallySN} is introduced. The main idea of the \textbf{sandwich rule} is that the performance of a large network should not be worse than that of a small network. Theoretically, the performances of all deep model with different channel configurations are bounded by the largest sub-network and the smallest sub-network. Therefore, as shown in Fig. \ref{fig:sub-networks}, we can train the deep model with $n$ sampled channel configurations in each training iteration, including the largest sub-network, the smallest sub-network, and $n-2$ randomly sampled sub-networks with different channel configurations. During training, we can explicitly track the validation performance of the largest model and the smallest model, which indicates the upper and lower bounds of the super network performance in theory. It is noteworthy that the trained super network is a dense neural network, which is compatible with existing GPUs and deep learning libraries. Specifically, the width multiplier $\rho$ is not uniformly applied to all layers in the super network. For a sub-network layer with the width multiplier $\rho$, the first $\left\lceil {\rho n} \right\rceil $ channels are activated where $\left\lceil  \cdot  \right\rceil $ denotes the ceiling operation.

To further improve the overall performance of the super network, we propose the \textbf{In-place Synergy} technique, which is inspired by the knowledge distillation techniques \cite{hinton2015distilling, tarvainen2017mean}. The main idea of In-place Synergy is to transfer knowledge obtained from a teacher network to a student network. Specifically, we use output of a larger sub-network to guide the training of the smaller sub-network. In each iteration, the largest sub-network is supervised by the ground truth image and the $n-2$ randomly sampled sub-networks are guided through the largest sub-network. As for the smallest sub-network, we use the ensemble of all sub-networks with different channel configurations. This is because the ensemble of multiple teacher networks ususaly generate more informative soft targets than a single teacher network \cite{tarvainen2017mean, you2017learning}. Formally, the total loss function for super network training can be defined as follows:
\begin{equation}
\mathcal{L} = D(\bm{\hat x}_L,\bm x) + D(\bm{\hat x}_R,\phi (\bm{\hat x}_L) + D(\bm{\hat x}_S,\phi (\bm{\hat x}_{R,L})) \\
\end{equation}
where $\bm{\hat{x}}_S = {\mathcal{F}_S}(\bm{\Theta};\bm{y} )$, $\bm{\hat{x}}_R = {\mathcal{F}_R}(\bm{\Theta};\bm{y} )$, $\bm{\hat{x}}_L = {\mathcal{F}_L}(\bm{\Theta};\bm{y} )$ are the predicted clean images by the smallest, randomly sampled, and the biggest sub-networks. $\bm x$, $\bm y$ represent the ground-truth and noisy image, respectively. $\phi(\cdot)$ denotes stop gradients, $D(\cdot,\cdot)$is a non-negative function that measures the divergence between the inputs. $\bm{\hat{x}}_{R,L}$ is the average of the clean images predicted by the randomly sampled and the largest sub-networks:
\begin{equation}
\bm{\hat{x}}_{R,L} = \frac{1}{{n - 1}}\Big( {{\mathcal{F}_L}(\bm{\Theta};\bm{y} ) + \sum\limits_{i = 1}^{n - 2} {{\mathcal{F}_R}(\bm{\Theta};\bm{y} )} } \Big).\\
\end{equation}

\subsection{Static Architecture Optimization via Progressive Slimming}
In the well-trained super network, the number of sub-networks with different channel configurations increases exponentially with the number of layers. Large numbers of sub-networks with different channel configurations have similar computation consumption, but a big difference in performance. Given a noisy input, it is extremely difficult to select a appropriate sub-network. Therefore, We attempt to reduce the search space of channel configurations and retain the sub-networks with the optimal channel configurations. Inspired by \cite{yu2019autoslim}, \textbf{progressive slimming} technique is presented to solve this problem.

\begin{figure}[!t]
\centerline{\includegraphics[width=0.45\textwidth]{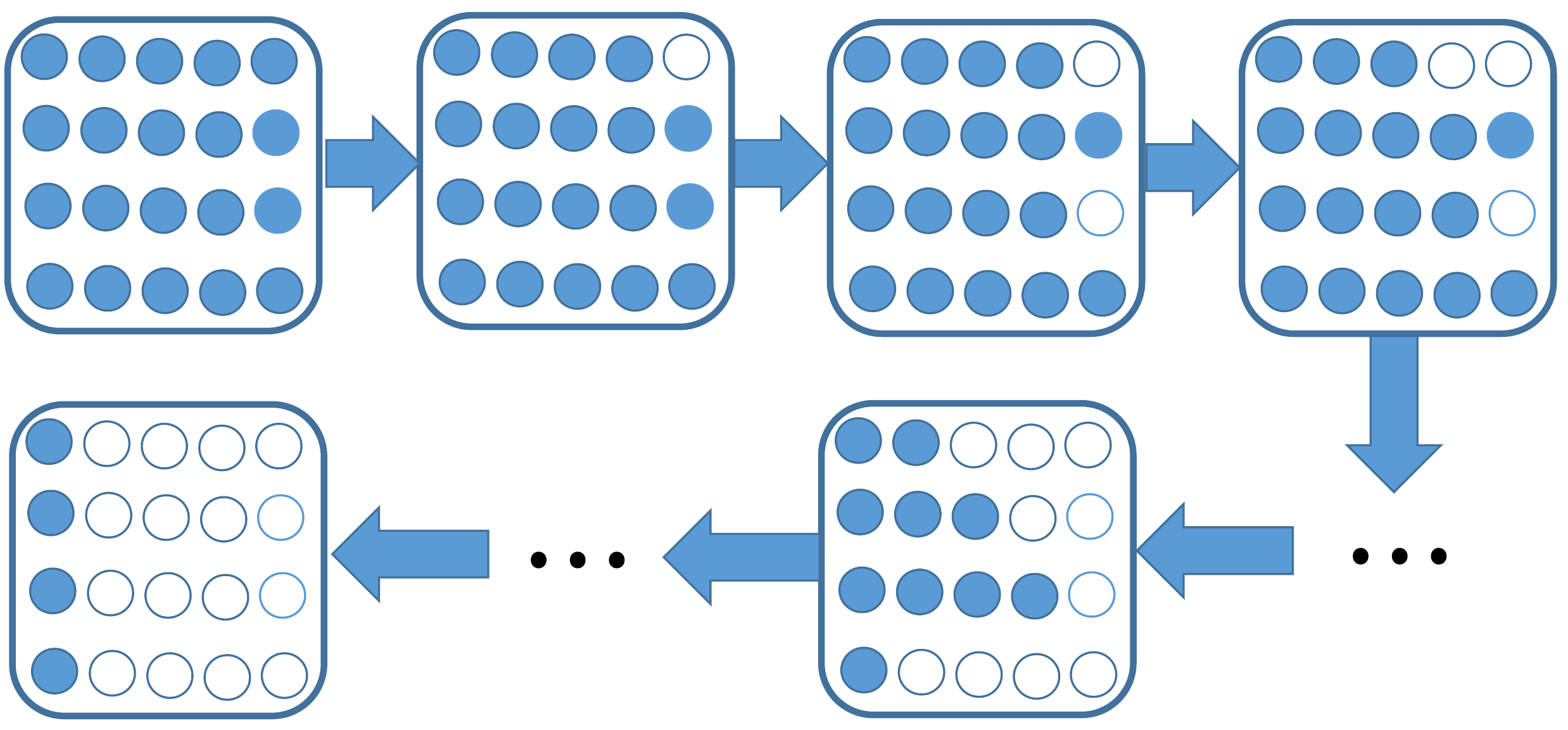}}
\caption{The pipeline of progressive slimming.}
\label{fig:progressive_slimming}
\end{figure}

The main idea of progressive slimming is to iteratively evaluate the network and progressively slim the layer with minimal performance drop. Fig. \ref{fig:progressive_slimming} illustrate the pipeline of progressive slimming. As displayed in Fig. \ref{fig:progressive_slimming}, given the well-trained super network, we iteratively slim the layers and evaluate the corresponding sub-network on the validation dataset. In each iteration, we can obtain $m$ sub-networks with diverse channel configurations, where $m$ is the number of super network layers. By comparing the performance of these sub-networks, we can decide to slim which layer in current iteration. In the next iterations, we repeat the evaluation-slim process until meeting the stop condition. Specifically, we start with the largest sub-network and end with the smallest sub-network. By a single pass, we can obtain a set of sub-networks, including the largest sub-network, the smallest sub-network and all the sub-networks with best performance. To further reduce the search space of the channel configurations, we divide the channels into several groups. Instead of an individual channel, we remove a group of channels during progressive slimming. For example, if a super network have $10$ layers and each layer have $64$ channels, there will be ${64^{10}}$ sub-networks before progressive slimming. After the progressive slimming, the number of sub-networks will be $64\times10{\rm{ = }}640$ and all these sub-networks have good performance. If the group number of each layer is $8$, the number of sub-networks will be further reduced to $64 \div 8 \times 10{\rm{ = 80}}$. In the experiments, we can change the number of channel groups to adjust the number of sub-networks.

\subsection{Dynamic Architecture Optimization via Dynamic gate}
In this section, we first design the dynamic gate function to select the appropriate sub-network conditioned on the noisy input images. Then, we present the training details of dynamic gate function.

\textbf{Gate function design.} 
After super network training and progressive slimming, a set of sub-networks with optimized channel configurations can be obtained. Given the noisy images, the gate function aims to make decisions on the selection of these sub-networks. The selection of sub-networks can be viewed as the image classification problem, where the sub-networks can be regarded as the label. There are many different ways of encoding the sub-networks, such as label encoding and one hot encoding. Label encoding gives each sub-network a number. It starts from $1$ and then increases for each sub-network. During training, neural network gives the higher weights to the higher numbers. Therefore, it has natural ordered relationships. Instead of label encoding, one hot encoding is adopted to ensure the fairness of the gate function. one hot encoding converts numbers to binary. It generates a one hot vector of ones and zeros. The length of one hot vector equals to the number of sub-networks with optimized channel configurations. 

\begin{figure}[!t]
\centerline{\includegraphics[width=0.4\textwidth]{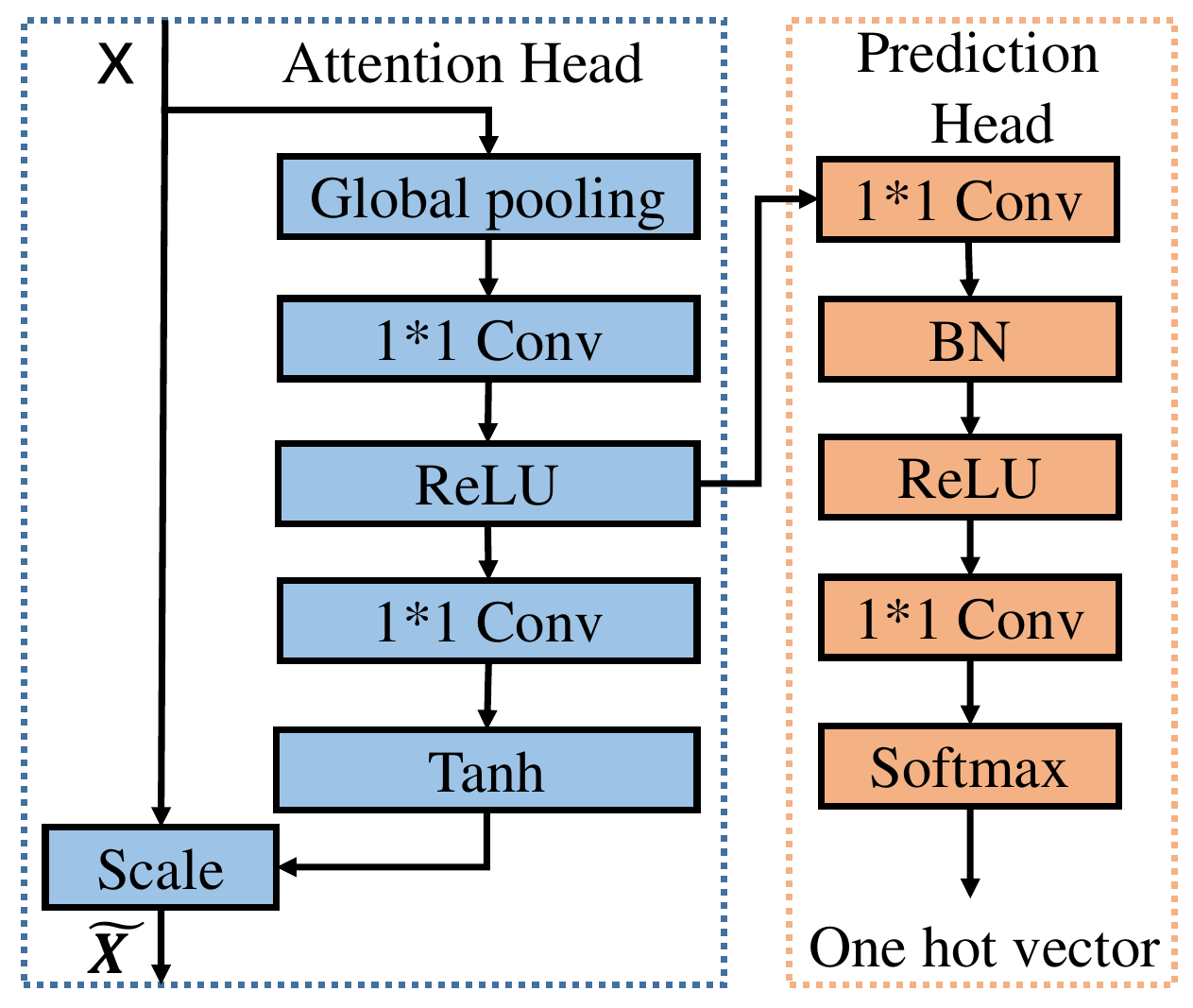}}
\caption{Overall architecture of the gate network.}
\label{fig:gate_architecture}
\end{figure}

Inspired by the recent channel attention and gating methods \cite{hu2018squeeze, yang2020gated, li2021dynamic}, we design a general and flexible gate network with one hot encoding, which can conveniently plug into any backbone network. Fig. \ref{fig:gate_architecture} displays a diagram of the gate network. As shown in Fig. \ref{fig:gate_architecture}, the gate function is composed of two components: an attention head and a prediction head. The attention head consists of two steps: squeeze and excitation.  Given an input feature map $\mathbf{X} \in {\mathbb{R}^{H \times W \times C}}$, the attention head first squeezes the global spatial information into a channel-wise statistic vector $U \in {\mathbb{R}^{1 \times 1 \times C}}$, which is calculated by:
\begin{equation}
U = \left\{ {\frac{1}{{H \times W}}\sum\limits_{i = 1}^H {\sum\limits_{j = 1}^W {{X_c}(i,j)} } } \right\}_{c = 1}^C.
\end{equation}
Then in the excitation step, a gating function with two $1*1$ convolutional layers and a tanh activation is employed to fully capture channel-wise dependencies. Formally, the attention head can be defined as follows: 
\begin{equation}
{\mathbf{\tilde X}}{\text{ = }}{\mathbf{X}}*\delta ({{\mathbf{W}}_2}(\sigma ({{\mathbf{W}}_1}U))),
\end{equation}
where ${{\mathbf{W}}_1} \in {\mathbb{R}^{\frac{C}{r} \times C}}$ is the parameters of dimensionality-reduction layer with reduction ratio $r$, ${{\mathbf{W}}_2} \in {\mathbb{R}^{C \times \frac{C}{r}}}$ is the parameters of dimensionality-increasing layer with reduction ratio $r$, $\sigma$ denotes the ReLU function, $\delta (x) = 1 + \tanh (x)$ refers to the activation function. As for the prediction head, it maps the reduced feature $R = \sigma ({{\mathbf{W}}_1}U)$ to a one hot vector $P$ used for the sub-network prediction. The prediction head is composed of two convolutional layers and a softmax layer. Specifically, the prediction head is defined as follows:
\begin{equation}
P = \arg \max \vartheta ({{\mathbf{W}}_4}(\sigma (BN({{\mathbf{W}}_3}R)))),
\end{equation}
where ${{\mathbf{W}}_3} \in {\mathbb{R}^{D \times \frac{C}{r}}}$ and ${{\mathbf{W}}_4} \in {\mathbb{R}^{E \times D}}$ are the parameters of two convolutional layers with kernel size $1$, D denotes the hidden dimension, E represents the number of candidate sub-networks, BN is the Batch Normalization and $\vartheta$ is the softmax function.

\textbf{Gate function training.}
Instance-wise dynamic networks adjusts their architectures conditioned on the inputs. To perform efficient and accurate dynamic inference, the dynamic networks should allocate less computation for easy inputs and more computation for hard inputs. To this end, the gate function of dynamic denoising networks should have the ability to identify easy and hard noisy images. 
Given the training noisy images, we first attempt to define their difficulty level and then generate the training label for them, which can be used to train the gate function.

As mentioned in \cite{Yu2019UniversallySN}, the performance of a cumbersome sub-network is always no worse than that of a compact sub-network. Therefore, we decide to define the difficulty level of noisy images according to the performance gains obtained by the cumbersome sub-network. Given a well-trained super network, we first evaluate the performance (PSNR) gain of the training noisy images with the largest sub-network and the smallest sub-network. Then the performance gain constraint is defined as follows:
\begin{equation}
\textit{\textbf{PSNR}}^{\bm +}(\bm y) = \textit{\textbf{PSNR}}_{l}(\bm y) - \textit{\textbf{PSNR}}_{s}(\bm y) > \beta,
\label{Eq:performance_gain}
\end{equation}
where $\textbf{PSNR}_{l}$ is the denoising results with the largest sub-network, $\textbf{PSNR}_{s}$ is the denosing results with the smallest sub-network and $\beta$ denotes a threshold. If the performance gain of a noisy image meets the equation \ref{Eq:performance_gain}, it can be viewed as a hard noisy image. Vice versa, it can be regard as the easy noisy image. In order to save the computation cost while maintaining the denoising performance, the gate function should allocate smaller sub-network for easy inputs and larger sub-network for hard inputs. Therefore, the gate label of training noisy images is generated as follows:

\begin{equation}
\mathbf{G}(\bm y) = \left\{ 
\begin{aligned}
  &{[1,0, \ldots ,0],{\textit{ if }\textit{\textbf{PSNR}}^{\bm +}(\bm y) < \beta}} \\ 
  &{[0,0, \ldots ,1],{\textit{ if }\textit{\textbf{PSNR}}^{\bm +}(\bm y) > \beta}} 
\end{aligned} \right.
\label{Eq:generate_online_label}
\end{equation}

In addition, the Cross-Entropy loss is introduced to measure the difference between the prediction of the gate network and the generated training label. Formally, the gate loss is formulated as follows:
\begin{equation}
{{\cal L}_{gate}}(\bm\varphi;\bm y) = {{\cal L}_{CE}}\big(\mathcal{G}(\bm\varphi; \bm y), \mathbf{G}(\bm y)\big)
\end{equation}
where $\bm y$ is the noisy image, $\mathcal{G}(\bm\varphi; \cdot)$ represents the gate network and $\mathbf{G}(\bm y)$ denotes the generated training label.

Another concern of the gate network training is to control the model complexity, which is important for a variety of resource-constrained platforms. Here, the FLOPs (number of multiply-adds) per pixel on the fly is adopted as the metrics of the model complexity. Specifically, the complexity loss is defined as follows:

\begin{equation}
{{\cal L}_{comp}}(\bm y; \bm\varphi) = {\left( {\frac{{\mathbf{\Psi}(\bm y,\bm\varphi ) - C}}{T}} \right)^2},
\end{equation}
where $\bm\varphi $ denotes the parameters of the gate function, $\mathbf{\Psi}(\bm y,\bm\varphi ) = \frac{{\textit{FLOPs}(\bm y,\bm\varphi )}}{{\textit{TotalPixel}}}$ is the FLOPs/Pixel for recovering $\bm x$ from $\bm y$, $C$ is the computation constraint of hardware, $T$ represents the normalization factor which is set to the FLOPs/Pixel of the super network in this paper.

Overall, the gate function be optimized with a joint function:
\begin{equation}
{\cal L} = {{\cal L}_{gate}} + {{\cal L}_{comp}},
\label{Eq:gate_loss}
\end{equation}
where ${\cal L}_{gate}$ is the gate loss and  ${\cal L}_{comp}$ is the complexity loss. Besides, it is noteworthy that the attention head is activated during the super network training and disabled during the gate network training. 

\section{Experiments}
\subsection{Settings}
\subsubsection{Dataset} 
Our DDS-Net is trained on SIDD-Medium dataset \cite{abdelhamed2018high}, which contains 320 image pairs (noisy and ground-truth) from 10 different scenes. These image pairs are captured under different lighting conditions using five representative smartphone cameras. To evaluate the performance of DDS-Net, we choose SIDD \cite{abdelhamed2018high} and DND \cite{plotz2017benchmarking} benchmark datasets as test datasets. \textbf{DND} benchmark dataset consists of 1,000 patches with a size of $512\times 512$ extracted from 50 real noisy images. \textbf{SIDD} benchmark dataset is comprised of 1,280 patches with a size of $256 \times 256$ extracted from 40 real noisy images. 

\subsubsection{Architecture and training details}
To evaluate our method, we use three representative state-of-the-art denoising networks: VDNet \cite{yue2019variational}, DANet \cite{yue2020dual}, MPRNet\cite{Zamir2021MPRNet}. The proposed DDS-Net empower these denoising networks the ability of automatically adjusting their channel configurations conditioned on the real-world noisy images. In this paper, these dynamic slimmable denoising networks are named as DDS-VDNet, DDS-DANet and DDS-MPRNet, respectively. In DDS-Net, there are three stages: (i) once-for-all weight optimization via In-place Synergy, (ii) static architecture optimization via progressive slimming, (iii) dynamic architecture optimization via dynamic gate. Next, we will illustrate the implementation and training details.

\noindent \textbf{Once-for-all weight optimization.} 
During the super network training, the number of sampled channel configurations is set as $n = 4$. In DDS-VDNet, DDS-DANet and DDS-MPRNet, the width ratio of the randomly sampled sub-network is set as $\rho  \in [0.2,1.5]$. In our implementation, the width ratio is non uniformly applied to all layers, which means that each layer has its own width ratio. For channels of each layer, we multiply $\rho$ and then round the resulting channel. The smallest division of channel number for each layer is set as $8$. The well-trained super network can execute at arbitrary widths bounded by the smallest and largest widths. Other training settings of super network are following corresponding papers \cite{yue2019variational, yue2020dual, Zamir2021MPRNet} respectively (for example, learning rate scheduling, weight initialization, data augmentation, patch size, batch size, optimizer, training iterations). 

\noindent \textbf{progressive Slimming.}
In the stage of progressive slimming, we divide channels of all layers into several groups and each group contains $16$ channels. After progressive slimming, we can obtain a set of sub-networks with their channel configurations and computation costs (FLOPs/pixel). Theoretically, the performances of these sub-networks are bounded by that of the smallest and largest sub-networks.

\noindent \textbf{Dynamic architecture optimization.}
The dynamic gate is inserted behind the first convolutional block of the DDS-Net. Given the real-word noisy images, the dynamic gate is responsible for selecting appropriate sub-networks. The threshold in Eq. \ref{Eq:performance_gain} is set as $\beta = 0.2$. For the gate training, we use Adam optimizer with $5{e^{ - 5}}$ initial learning rate for a total batch size of $64$. The learning rate decays by factor $0.9$ after every epoch. The patch size of training input is set as $128$.

\subsection{Main results}
To validate the effectiveness of the proposed method, we apply our method to three representative state-of-the-art denoising networks: VDNet \cite{yue2019variational}, DANet \cite{yue2020dual}, MPRNet \cite{Zamir2021MPRNet}. These weight-shared dynamic denoising networks are named as DDS-VDNet, DDS-DANet and DDS-MPRNet, respectively. We compare the DDS-Nets with the individual static denoising networks (I-Nets) at different computation resource constraints (FLOPs per pixel). Here, the runtime FLOPs per pixel is used to measure the efficiency of the denoising methods, while PSNR and SSIM are adopted to measure the denoising quality of them.

\begin{figure*}[!ht]
\centerline{\includegraphics[width=.9\textwidth]{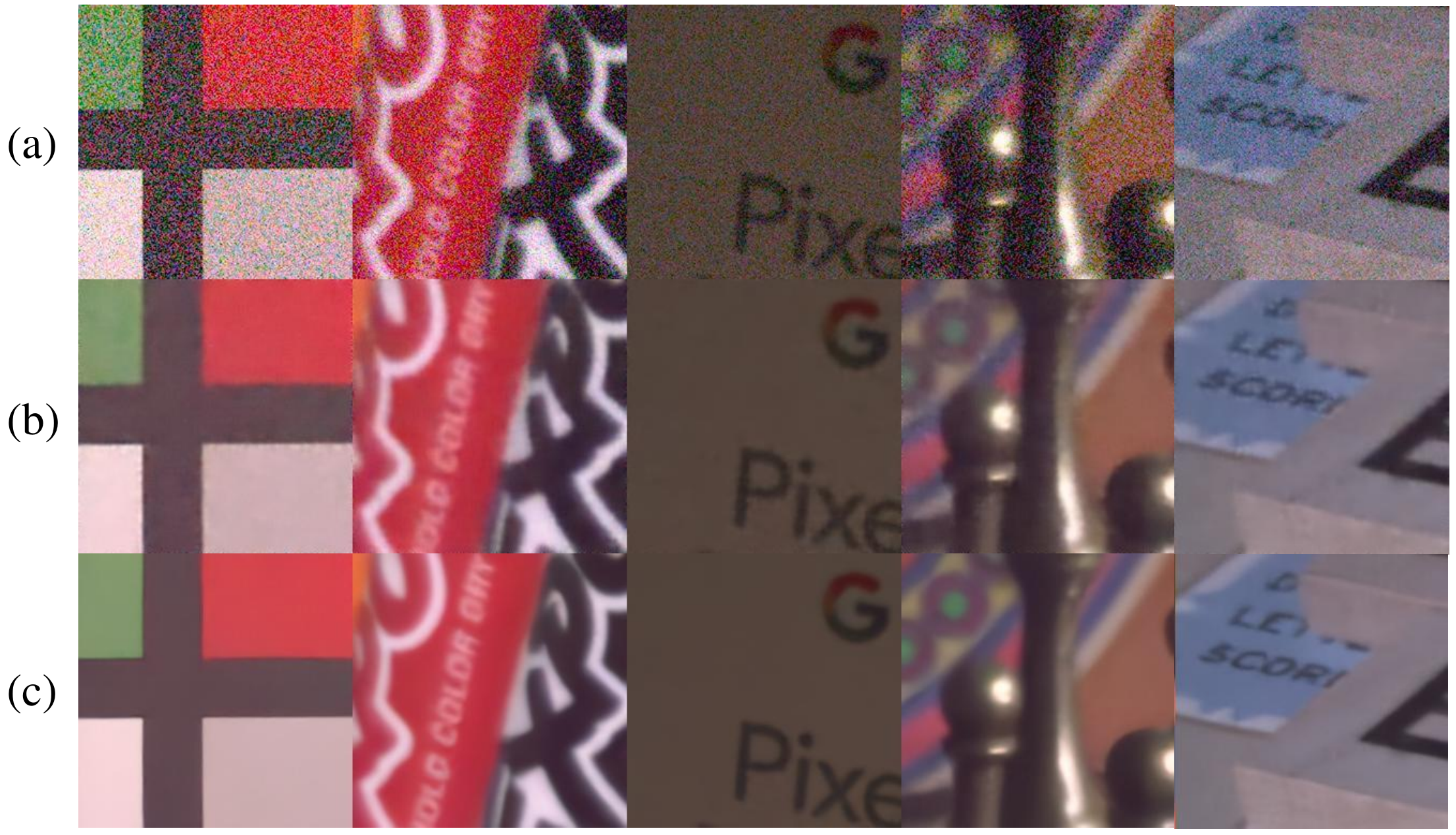}}
\caption{Denoising results of I-VDNet and DDS-VDNet on SIDD benchmank dataset when the computation constraint is 20K FLOPs/Pixel. (a) Noisy images, (b) Denoised images of I-VDNet, (c) Denoised images of DDS-VDNet.}
\label{fig:VDNet_flops_20_SIDD}
\end{figure*}

\begin{figure*}[!ht]
\centerline{\includegraphics[width=0.9\textwidth]{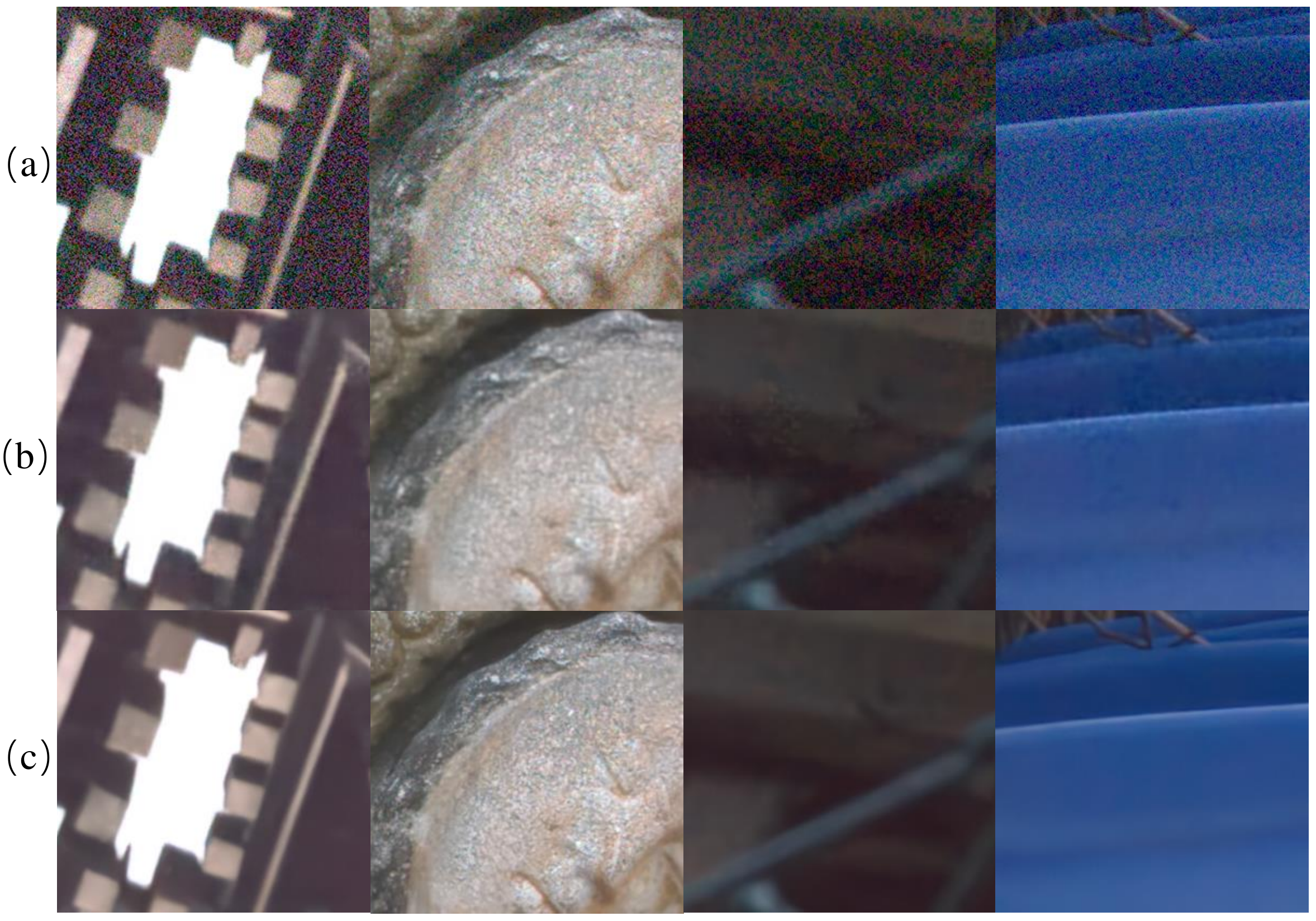}}
\caption{Denoising results of I-VDNet and DDS-VDNet on DND benchmank dataset when the computation constraint is 20K FLOPs/Pixel. (a) Noisy images, (b) Denoised images of I-VDNet, (c) Denoised images of DDS-VDNet.}
\label{fig:VDNet_flops_20_DND}
\end{figure*}

Table \ref{table:dynamic_compare_SIDD} and Table \ref{table:dynamic_compare_DND} summarize the comparison results of the static and dynamic denoising networks on SIDD and DND benchmark datasets. To have a clear view, we divide the network architectures into three groups for VDNet and DANet, namely, 20K FLOPs/Pixel, 100K FLOPs/Pixel and 200K FLOPs/Pixel. As MPRNet is a heavy network, we compare the static and dynamic versions at 500K FLOPs/Pixel, 1M FLOPs/Pixel and 5M FLOPs/Pixel. From Table \ref{table:dynamic_compare_SIDD} and Table \ref{table:dynamic_compare_DND}, we have the following observations: 1) with the increase of computation resource constraints (FLOPs/Pixel), both static and dynamic denoising networks achieve better performance; 2) the denoising performance of DDS-Nets consistently outperform I-Nets with almost the same computational cost under all computation resource constraints groups; 3) Under limited computation resource, such as 20K FLOPs/Pixel and 100K FLOPs/Pixel, DDS-Nets surpasses I-Nets by a large margin. We conjecture that the improvements may come from implicit knowledge distilling where the large model transfers its knowledge to small model by weight sharing and joint training. We also show some visual comparisons on SIDD and DND benchmark datasets in Fig. \ref{fig:VDNet_flops_20_SIDD} and Fig. \ref{fig:VDNet_flops_20_DND}. It can be seen that I-Nets generates artifacts and the denoising quality is unsatisfactory. In contrast, DDS-Nets can recover more fine details and avoid artifacts, resulting in good denoised images.


\begin{table}[t]
\renewcommand\arraystretch{1.2}
\footnotesize
\centering
\caption{Comparison results of various network architectures on SIDD benchmark dataset. I-Net denotes the static denoising model trained individually, while DDS-Net represents our dynamic weight-shared denoising model. {\color{red} Red} indicates our results.}
{\tabcolsep0.03in                     
\begin{tabular}{c|ccccccccc}
\toprule[1pt]
Group & Model & Type & FLOPs/Pixel & PSNR & SSIM  \\
\hline
\multirow{2}{*}{20K FLOPs/Pixel}
& I-VDNet & Static  &   16.8K & 37.89 & 0.943 \\
& \color{red}DDS-VDNet & Dynamic & 19.4K & 39.01 & 0.953\\
\hline
\multirow{2}{*}{100K FLOPs/Pixel}
& I-VDNet & Static  &   101K & 38.81 & 0.951 \\
& \color{red}DDS-VDNet & Dynamic & 113K & 39.24 & 0.955\\
\hline
\multirow{2}{*}{200K FLOPs/Pixel}
& I-VDNet & Static  &   205K & 39.04 & 0.953 \\
& \color{red}DDS-VDNet & Dynamic & 213K & 39.32 & 0.956\\
\hline
\multirow{2}{*}{20K FLOPs/Pixel}
& I-DANet & Static  &   24.3K & 38.93 & 0.952 \\
& \color{red}DDS-DANet & Dynamic & 22.4K & 39.11 & 0.954\\
\hline
\multirow{2}{*}{100K FLOPs/Pixel}
& I-DANet & Static  &   97.7K & 39.34 & 0.956 \\
& \color{red}DDS-DANet & Dynamic & 101K & 39.46 & 0.957\\
\hline
\multirow{2}{*}{200K FLOPs/Pixel}
& I-DANet & Static  &   199K & 39.42 & 0.957 \\
& \color{red}DDS-DANet & Dynamic & 207K & 39.54 & 0.958\\
\hline
\multirow{2}{*}{500K FLOPs/Pixel}
& I-MPRNet & Static  &   506K & 39.03 & 0.954 \\
& \color{red}DDS-MPRNet & Dynamic & 485K & 39.29 & 0.956\\
\hline
\multirow{2}{*}{1M FLOPs/Pixel}
& I-MPRNet & Static  &   0.97M & 39.36 & 0.956 \\
& \color{red}DDS-MPRNet & Dynamic & 0.94M & 39.43 & 0.957\\
\hline
\multirow{2}{*}{5M FLOPs/Pixel}
& I-MPRNet & Static  &   5.39M & 39.68 & 0.958 \\
& \color{red}DDS-MPRNet & Dynamic & 5.11M & 39.79 & 0.959\\
\bottomrule [1pt]
\end{tabular}
}
\label{table:dynamic_compare_SIDD}
\end{table}

\begin{table}[t]
\renewcommand\arraystretch{1.2}
\footnotesize
\centering
\caption{Comparison results of various network architectures on DND benchmark dataset. I-Net denotes the static denoising model trained individually, while DDS-Net represents our dynamic weight-shared denoising model. {\color{red} Red} indicates our results.}
{\tabcolsep0.03in                     
\begin{tabular}{c|ccccccccc}
\toprule[1pt]
Group & Model & Type & FLOPs/Pixel & PSNR & SSIM  \\
\hline
\multirow{2}{*}{20K FLOPs/Pixel}
& I-VDNet & Static  &   16.8K & 38.56 & 0.9437 \\
& \color{red}DDS-VDNet & Dynamic & 20.2K & 39.20 & 0.9485\\
\hline
\multirow{2}{*}{100K FLOPs/Pixel}
& I-VDNet & Static  &   101K & 39.08 & 0.9482 \\
& \color{red}DDS-VDNet & Dynamic & 106K & 39.33 & 0.9497 \\
\hline
\multirow{2}{*}{200K FLOPs/Pixel}
& I-VDNet & Static  &   205K & 39.18 & 0.9484 \\
& \color{red}DDS-VDNet & Dynamic & 197K & 39.41 & 0.9527\\
\hline
\multirow{2}{*}{20K FLOPs/Pixel}
& I-DANet & Static  &   24.3K & 39.06 & 0.9472 \\
& \color{red}DDS-DANet & Dynamic & 23.6K & 39.25 & 0.9493\\
\hline
\multirow{2}{*}{100K FLOPs/Pixel}
& I-DANet & Static  &   97.7K & 39.30 & 0.9499 \\
& \color{red}DDS-DANet & Dynamic & 104K & 39.46 & 0.9519\\
\hline
\multirow{2}{*}{200K FLOPs/Pixel}
& I-DANet & Static  &   199K & 39.38 & 0.9506 \\
& \color{red}DDS-DANet & Dynamic & 201K & 39.55 & 0.9525\\
\hline
\multirow{2}{*}{500K FLOPs/Pixel}
& I-MPRNet & Static  &   506K & 39.19 & 0.9589 \\
& \color{red}DDS-MPRNet & Dynamic & 499K & 39.37 & 0.9501\\
\hline
\multirow{2}{*}{1M FLOPs/Pixel}
& I-MPRNet & Static  &   0.97M & 39.45 & 0.9542 \\
& \color{red}DDS-MPRNet & Dynamic & 1.12M & 39.57 & 0.9546\\
\hline
\multirow{2}{*}{5M FLOPs/Pixel}
& I-MPRNet & Static  &   5.39M & 39.74 & 0.9549 \\
& \color{red}DDS-MPRNet & Dynamic & 5.03M & 39.92 & 0.9551\\
\bottomrule [1pt]
\end{tabular}
}
\label{table:dynamic_compare_DND}
\end{table}

\subsection{Ablation study}
\textbf{Knowledge distillation.}
Knowledge distillation plays an important role for the training of super network, which can transfer knowledge obtained from a cumbersome network to a compact network. In this paper, we use in-place synergy technique to assist the training of super network. To statistically validate the efficacy of the in-place synergy technique, we first train the super network with three different strategies: without in-place distillation (WoID), original in-place distillation (ID) and in-place synergy (IS). Then, we compare the denoising performance of the super network on its smallest and largest sub-networks for the SIDD benchmark dataset. Here, PSNR is used as the evaluation metric.

Table \ref{table:knowledge_distillation} shows the comparison results of these strategies with different denoising networks: VDNet, DANet and MPRNet. As displayed in Table \ref{table:knowledge_distillation}, the smallest and largest sub-networks trained with in-place synergy (IS) consistently obtain the best performance. With in-place distillation (ID), the VDNet super network can surpass baseline (WoID) 0.08dB and 0.01dB on its smallest and largest sub-networks. This is because knowledge distillation can transfer knowledge from largest sub-network to smallest sub-network by weight sharing and joint training. In contrast to in-place distillation, the in-place synergy can further boost the overall performance of the super network. This is because the ensemble of multiple networks usually generate more informative soft targets than a single network.

\begin{table}[t]
\renewcommand\arraystretch{1.2}
\footnotesize
\centering
\caption{Ablation analysis of the knowledge distillation technique with different architectures on the SIDD dataset.}
{\tabcolsep0.13in                     
\begin{tabular}{c|cccc}
\toprule[1pt]
Architecture & sub-network & WoID & ID & IS  \\
\hline
\multirow{2}{*}{VDNet}
 & smallest & 38.65 & 38.73 & 38.88 \\
 & largest & 39.27 & 39.28 & 39.34 \\
\hline
\multirow{2}{*}{DANet}
 & smallest & 38.77 & 38.84 & 39.02 \\
 & largest & 39.49 & 39.51 & 39.56 \\
\hline
\multirow{2}{*}{MPRNet}
 & smallest & 39.11 & 39.16 & 39.25 \\
 & largest & 39.74 & 39.75 & 39.82 \\
\bottomrule [1pt]
\end{tabular}
}
\label{table:knowledge_distillation}
\end{table}

\textbf{Progressive Slimming.}
In the well-trained super network, there are large numbers of sub-networks with different channel configurations. Progressive slimming is able to reduce the search space of channel configurations and remain the sub-networks with the optimal channel configurations. To validate the effect of progressive slimming, we compare the proposed method with the variant without progressive slimming, named baseline. To reduce the search space of sub-networks, we sample $50$ sub-networks with different channel configurations for the baseline method, including the largest sub-network, the smallest sub-network and $48$ randomly sampled sub-networks with different channel configurations. 

Table \ref{table:progressive_slimming} shows the ablation analysis of progressive slimming technique on the SIDD dataset in terms of PSNR and SSIM. As shown in Table \ref{table:progressive_slimming}, the baseline method obtain worse results than DDS-DANet. This is because the randomly sampled sub-networks are not optimal.

\begin{table}[!htbp]
\renewcommand\arraystretch{1.2}
\footnotesize
\centering
\caption{Ablation analysis of the progressive slimming on the SIDD dataset.}
{\tabcolsep0.07in                     
\begin{tabular}{c|cccc}
\toprule[1pt]
Groups & Methods & FLOPs/Pixel & PSNR & SSIM  \\
\hline
\multirow{2}{*}{20K FLOPs/Pixel}
& baseline   & 21.6K & 38.98 & 0.952 \\
& DDS-DANet  & 22.4K & 39.11 & 0.954\\
\hline
\multirow{2}{*}{100K FLOPs/Pixel}
& baseline   &  115K & 39.30 & 0.955 \\
& DDS-DANet  & 101K & 39.46 & 0.957\\
\hline
\multirow{2}{*}{200K FLOPs/Pixel}
& baseline   &  193K & 39.36 & 0.956 \\
& DDS-DANet & 207K & 39.54 & 0.958\\
\bottomrule [1pt]
\end{tabular}
}
\label{table:progressive_slimming}
\end{table}

\textbf{Effect of gate training.}
There are two losses used for the gate function training: gate loss ${\cal L}_{gate}$ and complexity loss ${\cal L}_{comp}$. With gate loss ${\cal L}_{gate}$, the dynamic model can adapt their architectures conditioned on the input noisy images. The main effect of the complexity loss ${\cal L}_{comp}$ is to control the computation complexity of the dynamic denoising model. Here, we perform extensive experiments with DANet to examine the impact of the gate loss. Table \ref{table:gate_loss} displays the ablation analysis of the gate loss on the SIDD dataset at three different computation resource constraints. As displayed in Table \ref{table:gate_loss}, the method with the gate loss consistently achieve better results. For input noisy images, the gate obtain its difficulty level. Then, the dynamic network allocate compact sub-network to the easy noisy images and cumbersome sub-network to the hard noisy images. 

\begin{table}[!htbp]
\renewcommand\arraystretch{1.2}
\footnotesize
\centering
\caption{Ablation analysis of the gate loss on the SIDD dataset.}
{\tabcolsep0.07in                     
\begin{tabular}{c|cccc}
\toprule[1pt]
Groups & gate loss & FLOPs/Pixel & PSNR & SSIM  \\
\hline
\multirow{2}{*}{20K FLOPs/Pixel}
& $\times$   & 21.6K & 39.04 & 0.953 \\
& \checkmark  & 22.4K & 39.11 & 0.954\\
\hline
\multirow{2}{*}{100K FLOPs/Pixel}
& $\times$    &  115K & 39.39 & 0.956 \\
& \checkmark  & 101K & 39.46 & 0.957\\
\hline
\multirow{2}{*}{200K FLOPs/Pixel}
& $\times$    &  193K & 39.48 & 0.958 \\
& \checkmark & 207K & 39.54 & 0.958\\
\bottomrule [1pt]
\end{tabular}
}
\label{table:gate_loss}
\end{table}

\textbf{Number of sampled channel configurations $n$.}
The number of sampled channel configurations $n$ per iteration is of central importance for the super network training. Intuitively, large $n$ leads to more GPU memory cost and training time. To analyze the effect of $n$, we train the super networks with $n = 3,4,5$. Table \ref{table:channel_configurations} summarizes the comparison results of DDS-Nets on the SIDD dataset with respect to different $n$. Here, we compare the performance of the super networks on its smallest and largest sub-networks. From Table \ref{table:channel_configurations}, we have the following observations: 1) compared to the super networks trained with $n=4, 5$, the super networks trained with $n = 3$ achieves the worse results; 2) the super network trained with $n=4$ shares similar performance with that trained with $n=5$. As large $n$ needs more resources, we set $n=4$ in our experiments.    

\begin{table}[!htbp]
\renewcommand\arraystretch{1.2}
\footnotesize
\centering
\caption{Experimental results of DDS-Nets on SIDD dataset regarding different number of sampled channel configurations $n$.}
{\tabcolsep0.13in                     
\begin{tabular}{c|cccc}
\toprule[1pt]
Architecture & sub-network & $n=3$ & $n=4$ & $n=5$  \\
\hline
\multirow{2}{*}{VDNet}
 & smallest & 38.81  & 38.88 & 39.89 \\
 & largest & 39.31  & 39.34 & 39.32\\
\hline
\multirow{2}{*}{DANet}
 & smallest & 39.01  & 39.02 & 39.02\\
 & largest & 39.49  & 39.56 & 39.57\\
\hline
\multirow{2}{*}{MPRNet}
 & smallest & 39.22  & 39.25 & 39.26\\
 & largest & 39.76  & 39.82 & 39.80\\
\bottomrule [1pt]
\end{tabular}
}
\label{table:channel_configurations}
\end{table}

\begin{figure}[!ht]
\centerline{\includegraphics[width=0.4\textwidth]{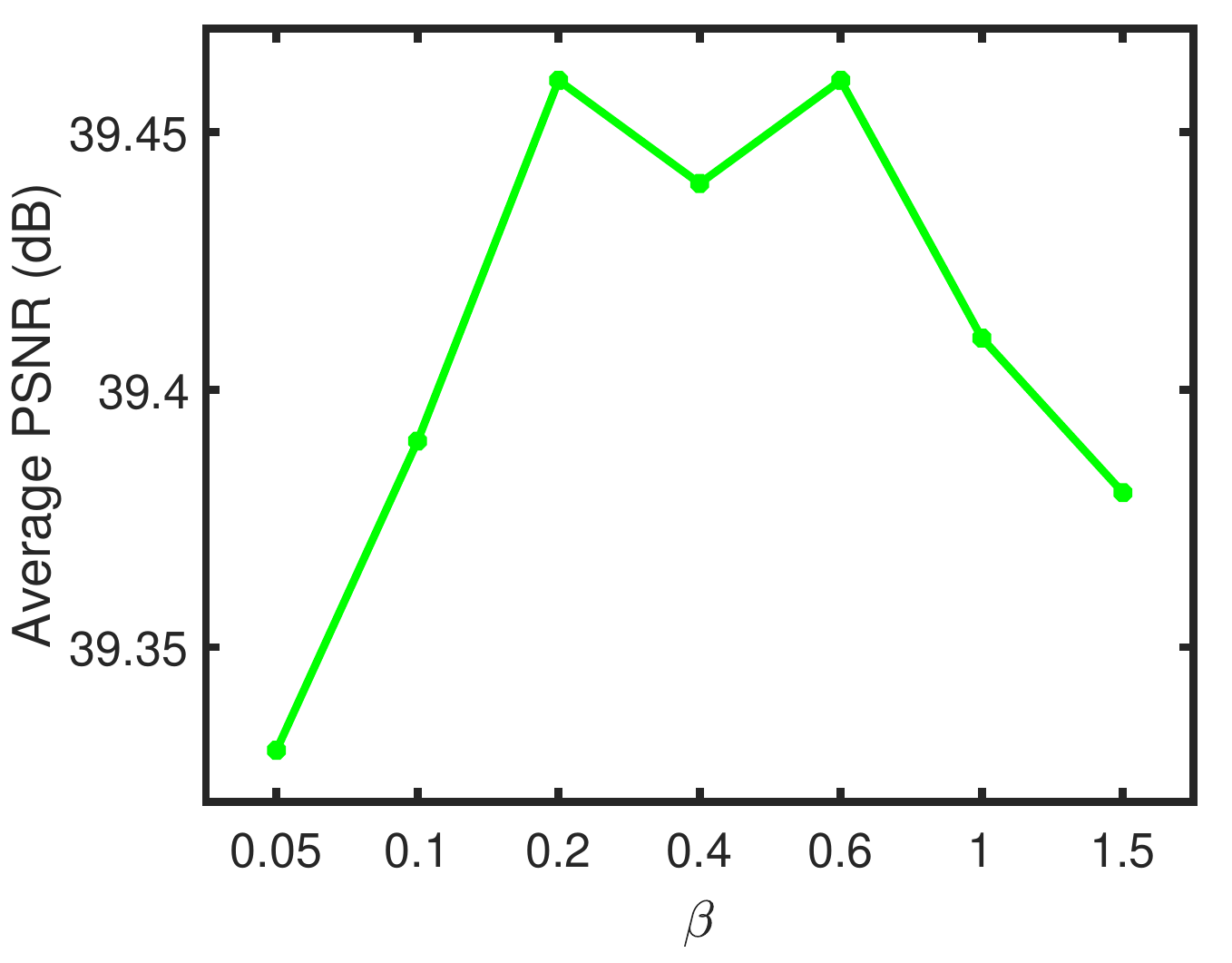}}
\caption{Experimental results of DDS-DANet on SIDD dataset with respect to $\beta$.}
\label{fig:effect_beta}
\end{figure}

\textbf{Hyper parameter $\beta$ analysis.} In Eq. \ref{Eq:performance_gain}, we define a threshold $\beta$ to identify the difficulty level of the training noisy images. To investigate how the threshold $\beta$ affects the training of gate function, we conduct an ablation analysis with the architecture DANet on the SIDD dataset. Here, we fix the computation constraint as $C=100K$ FLOPs/Pixel and vary the threshold $\beta$. As displayed in Fig. \ref{fig:effect_beta}, the models trained with $\beta = 0.2, 0.6$ obtain the best results. By adjusting the threshold $\beta$. we can adjust the ratio between the hard inputs and easy inputs. Finally, we can achieve good trade-off between denoising quality and computation complexity.

\vspace{-.8 em}
\section{Conclusion}
In this paper, we propose an general method to achieve good trade-off between denoising quality and computation complexity. The proposed method, named DDS-Nets, can adjust its channel configurations conditioned on the input noisy images at test time. Specifically, DDS-Nets allocate the cumbersome sub-network to the hard noisy images and compact sub-network to the easy noisy images. Extensive experiments demonstrate our DDS-Net consistently outperforms the state-of-the-art individually trained static denoising networks.

\section*{Acknowledgment}
Zutao Jiang was partially supported by China Scholarship Council. Dr Xiaojun Chang was partially supported by Australian Research Council (ARC) Discovery Early Career Researcher Award (DECRA) under grant no. DE190100626.

\begin{bibliographystyle}{IEEEtran}
\begin{bibliography}{mybib}
\end{bibliography}
\end{bibliographystyle}



\vspace{-1.8 em}
\begin{IEEEbiography}[{\includegraphics[width=1in,height=1.25in,clip,keepaspectratio]{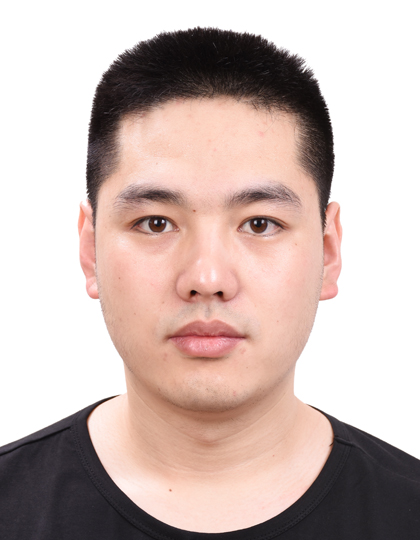}}]
{Zutao Jiang} received his B.S. degree from software engineering, Central South University, China in 2015. He is pursuing the Ph.D degree in the School of Software Engineering, Xi’an Jiaotong University. He was awarded a Chinese Government Scholarship by the China Scholarship Council to study at Monash University from September 2019 to September 2021. His research interests include computer vision and machine learning.
\end{IEEEbiography}

\vspace{-1.8 em}
\begin{IEEEbiography}[{\includegraphics[width=1in,height=1.25in,clip,keepaspectratio]{Authors/ChanglinLi}}]{Changlin Li} is a Ph.D. candidate in GORSE Lab, Faculty of Information Technology at Monash University, Melbourne Australia. Prior to arriving at Monash, he received his B.E. degree in Computer Science in 2019, from University of Science and Technology of China. He currently serves as a reviewer of ICCV and T-IP. His main research ambition is to explore more efficient and more intelligent neural network architectures. He also interests in computer vision tasks such like activity recognition and has won the first place in the TRECVID ActEV 2019 grand challenge.\end{IEEEbiography}

\vspace{-1.8 em}
\begin{IEEEbiography}[{\includegraphics[width=1in,height=1.25in,clip,keepaspectratio]{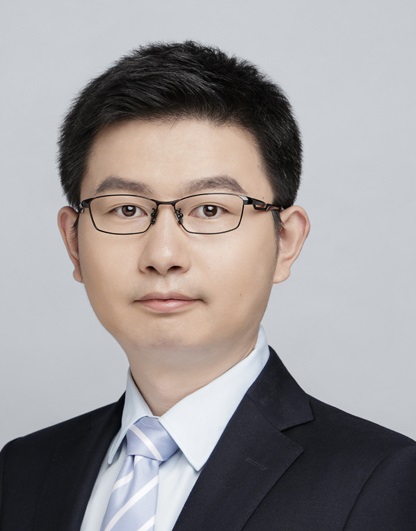}}]
{Xiaojun Chang} is an Associate Professor in School of Computing Technologies, RMIT University, Australia. He is an ARC Discovery Early Career Researcher Award (DECRA) Fellow between 2019 and 2021. Before joining RMIT, he was a Senior Lecturer in Faculty of Information Technology, Monash University from December 2018 to July 2021. Before joining Monash, he was a Postdoc Research Associate in School of Computer Science, Carnegie Mellon University. He has spent most of time working on exploring multiple signals (visual, acoustic, textual) for automatic content analysis in unconstrained or surveillance videos. He has achieved top performance in various international competitions, such as TRECVID MED, TRECVID SIN, and TRECVID AVS.
\end{IEEEbiography}

\vspace{-1.8 em}
\begin{IEEEbiography}[{\includegraphics[width=1in,height=1.25in,clip,keepaspectratio]{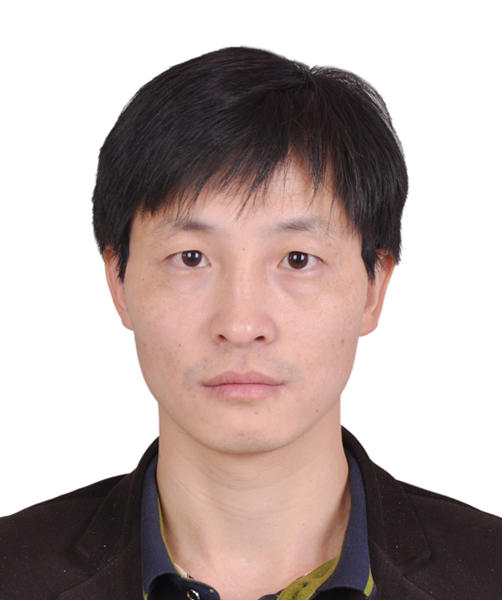}}]
{Jihua Zhu} received the B.E. degree in automation from Central South University, China, in 2004, respectively, and the Ph.D. degree in pattern recognition and intelligence systems from Xi’an Jiaotong University, China, in 2011. He is currently an Associate Professor with the School of Software Engineering, Xi’an Jiaotong University. His research interests include computer vision and machine learning.
\end{IEEEbiography}

\vspace{-1.8 em}
\begin{IEEEbiography}[{\includegraphics[width=1in,height=1.25in,clip,keepaspectratio]{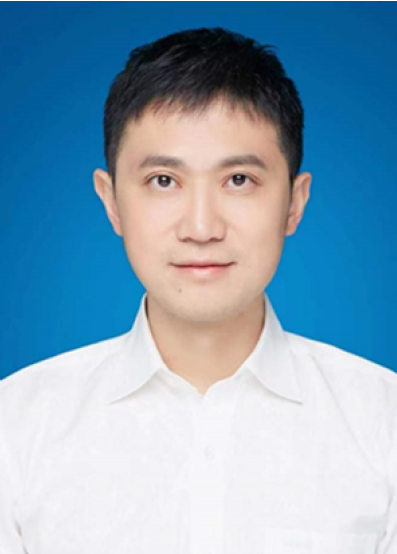}}]
{Yi Yang} (Senior Member, IEEE) received the Ph.D. degree in computer science from Zhejiang University, Hangzhou, China, in 2010. He was a Postdoctoral Researcher with the School of Computer Science, Carnegie Mellon University, Pittsburgh, PA, USA. He is currently a Professor with the University of Technology Sydney, Australia. His current research interests include machine learning and its applications to multimedia content analysis and computer vision, such as multimedia indexing and retrieval, surveillance video analysis, and video content understanding
\end{IEEEbiography}


\begin{thebibliography}{10}
\providecommand{\url}[1]{#1}
\csname url@samestyle\endcsname
\providecommand{\newblock}{\relax}
\providecommand{\bibinfo}[2]{#2}
\providecommand{\BIBentrySTDinterwordspacing}{\spaceskip=0pt\relax}
\providecommand{\BIBentryALTinterwordstretchfactor}{4}
\providecommand{\BIBentryALTinterwordspacing}{\spaceskip=\fontdimen2\font plus
\BIBentryALTinterwordstretchfactor\fontdimen3\font minus
  \fontdimen4\font\relax}
\providecommand{\BIBforeignlanguage}[2]{{%
\expandafter\ifx\csname l@#1\endcsname\relax
\typeout{** WARNING: IEEEtran.bst: No hyphenation pattern has been}%
\typeout{** loaded for the language `#1'. Using the pattern for}%
\typeout{** the default language instead.}%
\else
\language=\csname l@#1\endcsname
\fi
#2}}
\providecommand{\BIBdecl}{\relax}
\BIBdecl

\bibitem{howard2017mobilenets}
A.~G. Howard, M.~Zhu, B.~Chen, D.~Kalenichenko, W.~Wang, T.~Weyand,
  M.~Andreetto, and H.~Adam, ``Mobilenets: Efficient convolutional neural
  networks for mobile vision applications,'' \emph{Arxiv:1704.04861}, 2017.

\bibitem{sandler2018mobilenetv2}
M.~Sandler, A.~Howard, M.~Zhu, A.~Zhmoginov, and L.-C. Chen, ``Mobilenetv2:
  Inverted residuals and linear bottlenecks,'' in \emph{CVPR}, 2018, pp.
  4510--4520.

\bibitem{tan2019efficientnet}
M.~Tan and Q.~Le, ``Efficientnet: Rethinking model scaling for convolutional
  neural networks,'' in \emph{ICML}, 2019, pp. 6105--6114.

\bibitem{he2018soft}
Y.~He, G.~Kang, X.~Dong, Y.~Fu, and Y.~Yang, ``Soft filter pruning for
  accelerating deep convolutional neural networks,'' in \emph{IJCAI}, 2018.

\bibitem{he2017channel}
Y.~He, X.~Zhang, and J.~Sun, ``Channel pruning for accelerating very deep
  neural networks,'' in \emph{ICCV}, 2017, pp. 1389--1397.

\bibitem{liu2019metapruning}
Z.~Liu, H.~Mu, X.~Zhang, Z.~Guo, X.~Yang, K.-T. Cheng, and J.~Sun,
  ``Metapruning: Meta learning for automatic neural network channel pruning,''
  in \emph{ICCV}, 2019, pp. 3296--3305.

\bibitem{hinton2015distilling}
G.~Hinton, O.~Vinyals, and J.~Dean, ``Distilling the knowledge in a neural
  network,'' in \emph{NeurIPS Workshop}, 2014.

\bibitem{jacob2018quantization}
B.~Jacob, S.~Kligys, B.~Chen, M.~Zhu, M.~Tang, A.~Howard, H.~Adam, and
  D.~Kalenichenko, ``Quantization and training of neural networks for efficient
  integer-arithmetic-only inference,'' in \emph{CVPR}, 2018, pp. 2704--2713.

\bibitem{hua2018channel}
W.~Hua, Y.~Zhou, C.~De~Sa, Z.~Zhang, and G.~E. Suh, ``Channel gating neural
  networks,'' in \emph{NeurIPS}, vol.~32, 2019.

\bibitem{veit2018convolutional}
A.~Veit and S.~Belongie, ``Convolutional networks with adaptive inference
  graphs,'' in \emph{ECCV}, 2018, pp. 3--18.

\bibitem{wang2018skipnet}
X.~Wang, F.~Yu, Z.-Y. Dou, T.~Darrell, and J.~E. Gonzalez, ``Skipnet: Learning
  dynamic routing in convolutional networks,'' in \emph{ECCV}, 2018, pp.
  409--424.

\bibitem{li2021dynamic}
C.~Li, G.~Wang, B.~Wang, X.~Liang, Z.~Li, and X.~Chang, ``Dynamic slimmable
  network,'' in \emph{CVPR}, 2021, pp. 8607--8617.

\bibitem{gu2019self}
S.~Gu, Y.~Li, L.~V. Gool, and R.~Timofte, ``Self-guided network for fast image
  denoising,'' in \emph{ICCV}, 2019, pp. 2511--2520.

\bibitem{xu2021efficient}
L.~Xu, J.~Zhang, X.~Cheng, F.~Zhang, X.~Wei, and J.~Ren, ``Efficient deep image
  denoising via class specific convolution,'' in \emph{AAAI}, vol.~35, no.~4,
  2021, pp. 3039--3046.

\bibitem{han2021dynamic}
Y.~Han, G.~Huang, S.~Song, L.~Yang, H.~Wang, and Y.~Wang, ``Dynamic neural
  networks: A survey,'' \emph{ArXiv:2102.04906}, 2021.

\bibitem{huang2017multi}
G.~Huang, D.~Chen, T.~Li, F.~Wu, L.~van~der Maaten, and K.~Q. Weinberger,
  ``Multi-scale dense networks for resource efficient image classification,''
  in \emph{ICLR}, 2018.

\bibitem{yuan2020s2dnas}
Z.~Yuan, B.~Wu, G.~Sun, Z.~Liang, S.~Zhao, and W.~Bi, ``S2dnas: Transforming
  static cnn model for dynamic inference via neural architecture search,'' in
  \emph{ECCV}, 2020, pp. 175--192.

\bibitem{tanno2019adaptive}
R.~Tanno, K.~Arulkumaran, D.~Alexander, A.~Criminisi, and A.~Nori, ``Adaptive
  neural trees,'' in \emph{ICML}, 2019, pp. 6166--6175.

\bibitem{yu2018slimmable}
J.~Yu, L.~Yang, N.~Xu, J.~Yang, and T.~Huang, ``Slimmable neural networks,'' in
  \emph{ICLR}, 2019.

\bibitem{Yu2019UniversallySN}
J.~Yu and T.~S. Huang, ``Universally slimmable networks and improved training
  techniques,'' in \emph{ICCV}, 2019, pp. 1803--1811.

\bibitem{jain2008natural}
V.~Jain and S.~Seung, ``Natural image denoising with convolutional networks,''
  in \emph{NeurIPS}, vol.~21, 2008.

\bibitem{zhang2017beyond}
K.~Zhang, W.~Zuo, Y.~Chen, D.~Meng, and L.~Zhang, ``Beyond a gaussian denoiser:
  Residual learning of deep cnn for image denoising,'' \emph{TIP}, vol.~26,
  no.~7, pp. 3142--3155, 2017.

\bibitem{mao2016image}
X.~Mao, C.~Shen, and Y.-B. Yang, ``Image restoration using very deep
  convolutional encoder-decoder networks with symmetric skip connections,'' in
  \emph{NeurIPS}, vol.~29, 2016, pp. 2802--2810.

\bibitem{tai2017memnet}
Y.~Tai, J.~Yang, X.~Liu, and C.~Xu, ``Memnet: A persistent memory network for
  image restoration,'' in \emph{ICCV}, 2017, pp. 4539--4547.

\bibitem{liu2018non}
D.~Liu, B.~Wen, Y.~Fan, C.~C. Loy, and T.~S. Huang, ``Non-local recurrent
  network for image restoration,'' in \emph{NeurIPS}, 2018.

\bibitem{liu2018multi}
P.~Liu, H.~Zhang, K.~Zhang, L.~Lin, and W.~Zuo, ``Multi-level wavelet-cnn for
  image restoration,'' in \emph{CVPR workshops}, 2018, pp. 773--782.

\bibitem{guo2019toward}
S.~Guo, Z.~Yan, K.~Zhang, W.~Zuo, and L.~Zhang, ``Toward convolutional blind
  denoising of real photographsyu,'' in \emph{CVPR}, 2019, pp. 1712--1722.

\bibitem{anwar2019real}
S.~Anwar and N.~Barnes, ``Real image denoising with feature attention,'' in
  \emph{ICCV}, 2019, pp. 3155--3164.

\bibitem{yue2019variational}
Z.~Yue, H.~Yong, Q.~Zhao, L.~Zhang, and D.~Meng, ``Variational denoising
  network: Toward blind noise modeling and removal,'' in \emph{NeurIPS}, 2019.

\bibitem{yue2020dual}
Z.~Yue, Q.~Zhao, L.~Zhang, and D.~Meng, ``Dual adversarial network: Toward
  real-world noise removal and noise generation,'' in \emph{ECCV}, 2020, pp.
  41--58.

\bibitem{zamir2020learning}
S.~W. Zamir, A.~Arora, S.~Khan, M.~Hayat, F.~S. Khan, M.-H. Yang, and L.~Shao,
  ``Learning enriched features for real image restoration and enhancement,'' in
  \emph{ECCV}, 2020, pp. 492--511.

\bibitem{kim2020transfer}
Y.~Kim, J.~W. Soh, G.~Y. Park, and N.~I. Cho, ``Transfer learning from
  synthetic to real-noise denoising with adaptive instance normalization,'' in
  \emph{CVPR}, 2020, pp. 3482--3492.

\bibitem{Zamir2021MPRNet}
S.~W. Zamir, A.~Arora, S.~Khan, M.~Hayat, F.~S. Khan, M.-H. Yang, and L.~Shao,
  ``Multi-stage progressive image restoration,'' in \emph{CVPR}, 2021.

\bibitem{bolukbasi2017adaptive}
T.~Bolukbasi, J.~Wang, O.~Dekel, and V.~Saligrama, ``Adaptive neural networks
  for efficient inference,'' in \emph{ICML}, 2017, pp. 527--536.

\bibitem{lin2017runtime}
J.~Lin, Y.~Rao, J.~Lu, and J.~Zhou, ``Runtime neural pruning,'' in
  \emph{NeurIPS}, 2017, pp. 2178--2188.

\bibitem{liu2019learning}
C.~Liu, Y.~Wang, K.~Han, C.~Xu, and C.~Xu, ``Learning instance-wise sparsity
  for accelerating deep models,'' \emph{arXiv preprint arXiv:1907.11840}, 2019.

\bibitem{yu2019autoslim}
J.~Yu and T.~Huang, ``Autoslim: Towards one-shot architecture search for
  channel numbers,'' in \emph{NeurIPS Workshop}, 2019.

\bibitem{komodakis2017paying}
N.~Komodakis and S.~Zagoruyko, ``Paying more attention to attention: improving
  the performance of convolutional neural networks via attention transfer,'' in
  \emph{ICLR}, 2017.

\bibitem{tarvainen2017mean}
A.~Tarvainen and H.~Valpola, ``Mean teachers are better role models:
  Weight-averaged consistency targets improve semi-supervised deep learning
  results,'' in \emph{NeurIPS}, 2017.

\bibitem{you2017learning}
S.~You, C.~Xu, C.~Xu, and D.~Tao, ``Learning from multiple teacher networks,''
  in \emph{SIGKDD}, 2017, pp. 1285--1294.

\bibitem{hu2018squeeze}
J.~Hu, L.~Shen, and G.~Sun, ``Squeeze-and-excitation networks,'' in
  \emph{CVPR}, 2018, pp. 7132--7141.

\bibitem{yang2020gated}
Z.~Yang, L.~Zhu, Y.~Wu, and Y.~Yang, ``Gated channel transformation for visual
  recognition,'' in \emph{CVPR}, 2020, pp. 11\,794--11\,803.

\bibitem{abdelhamed2018high}
A.~Abdelhamed, S.~Lin, and M.~S. Brown, ``A high-quality denoising dataset for
  smartphone cameras,'' in \emph{CVPR}, 2018, pp. 1692--1700.

\bibitem{plotz2017benchmarking}
T.~Plotz and S.~Roth, ``Benchmarking denoising algorithms with real
  photographs,'' in \emph{CVPR}, 2017, pp. 1586--1595.

\end{thebibliography}
\end{document}